\documentclass[conference, letterpaper]{IEEEtran}
\IEEEoverridecommandlockouts

\usepackage{cite}
\usepackage{amsmath,amssymb,amsfonts}
\usepackage{algorithmic}
\usepackage{graphicx}
\usepackage{textcomp}
\usepackage{xcolor}
\usepackage[letterpaper, right=1.7cm, left=1.7cm, top=1.9cm, bottom=3.3cm]{geometry}
\def\BibTeX{{\rm B\kern-.05em{\sc i\kern-.025em b}\kern-.08em
    T\kern-.1667em\lower.7ex\hbox{E}\kern-.125emX}}

\begin{document}
\title{DVFS-Aware DNN Inference on GPUs:\\ Latency Modeling and Performance Analysis}

\author{\IEEEauthorblockN{
Yunchu~Han,~Zhaojun~Nan,~Sheng~Zhou,~and~Zhisheng~Niu
}
\thanks{This work is supported in part by the National Natural Science Foundation of China under Grants 62341108, in part by the China Postdoctoral Science Foundation under Grant 2023M742011, and in part by Hitachi Ltd.}
\IEEEauthorblockA{
Beijing National Research Center for Information Science and Technology\\Department of Electronic Engineering, Tsinghua University, Beijing 100084, China}

Emails: \{hyc23@mails.,~nzj660624@mail.,~sheng.zhou@,~niuzhs@\}tsinghua.edu.cn}

\maketitle

\begin{abstract}
The rapid development of deep neural networks (DNNs) is inherently accompanied by the problem of high computational costs. To tackle this challenge, dynamic voltage frequency scaling (DVFS) is emerging as a promising technology for balancing the latency and energy consumption of DNN inference by adjusting the computing frequency of processors. However, most existing models of DNN inference time are based on the CPU-DVFS technique, and directly applying the CPU-DVFS model to DNN inference on GPUs will lead to significant errors in optimizing latency and energy consumption. In this paper, we propose a DVFS-aware latency model to precisely characterize DNN inference time on GPUs. We first formulate the DNN inference time based on extensive experiment results for different devices and analyze the impact of fitting parameters. Then by dividing DNNs into multiple blocks and obtaining the actual inference time, the proposed model is further verified. Finally, we compare our proposed model with the CPU-DVFS model in two specific cases. Evaluation results demonstrate that local inference optimization with our proposed model achieves a reduction of no less than 66\% and 69\% in inference time and energy consumption respectively. In addition, cooperative inference with our proposed model can improve the partition policy and reduce the energy consumption compared to the CPU-DVFS model.
\end{abstract}

\section{Introduction}\label{sec:intro}
With the rapid advancement of artificial intelligence (AI), deep neural networks (DNNs) have gained increasing traction for real-time data processing and various applications, such as image recognition, object detection, speech recognition, and natural language processing \cite{Liu1}. The intricate structure and enormously multiple parameters of DNNs lead to high computational demands, rendering DNN inference both time-consuming and energy-intensive. Therefore, utilizing the limited computing capabilities of mobile devices is insufficient to fulfill the stringent quality of service (QoS) requirement. To mitigate this challenge, the concept of mobile edge computing (MEC) \cite{Mao2} has been proposed as a potential technology to provide real-time services at the wireless network edges (e.g., base stations). In MEC and edge intelligence \cite{Zhou3} systems, dynamic voltage and frequency scaling (DVFS) \cite{Wang4} technique is commonly used to balance the performance of processors by adjusting the computing frequency based on the real-time energy consumption. In addition, by fully or partially offloading DNN inference tasks from mobile devices to the edge servers (e.g., DNN partitioning), we can potentially reduce both inference time and energy consumption of mobile devices.

\begin{figure}[t]
  \centering
  \includegraphics[width=0.32\textwidth]{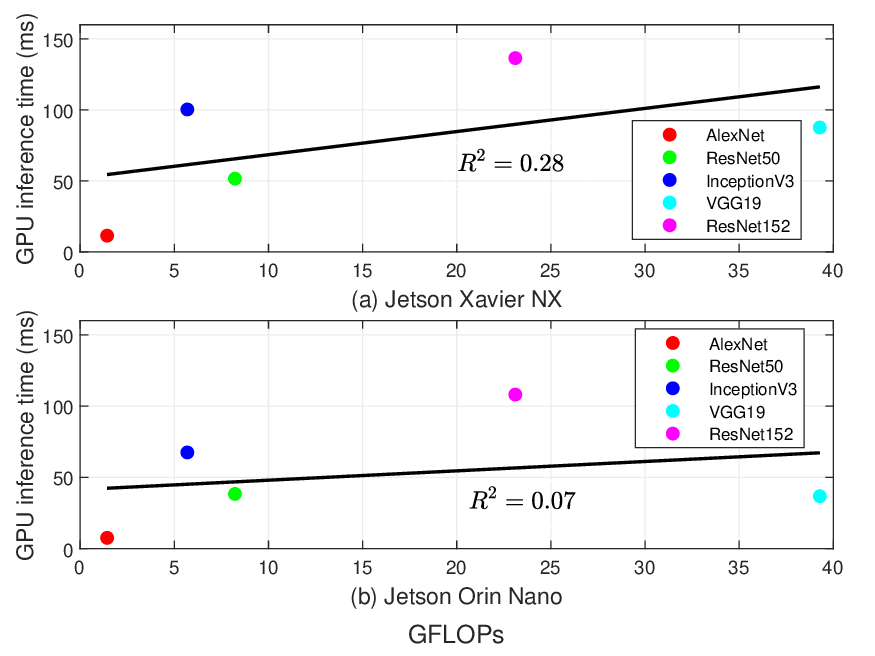}\\
  \caption{Average GPU inference time of different DNNs on (a) Jetson Xavier NX and (b) Jetson Orin Nano when GPU frequency is fixed.}
  \label{fig:myFig1}
\end{figure}

In order to enhance the efficiency of inference tasks in terms of computing frequency regulation, a precise latency model for DNN inference is essential. A common method in MEC is based on the CPU-DVFS model, which represents the execution time of a task as the ratio of required CPU cycles to CPU frequency \cite{Fan5}. Based on this model, joint optimization of task offloading and resource allocation is taken into account in many scenarios (e.g., vehicular edge computing \cite{Nan6} and unmanned aerial vehicle (UAV)-enabled edge computing \cite{Zhao7}). In \cite{Zeng8}, the authors have leveraged this model to characterize the DNN inference time on CPUs. However, the number of CPU cycles DNN inference needs is uncertain and obtained by estimation, which may have an impact on offloading policy, latency and energy consumption \cite{Nan9}. Besides, it becomes more and more popular to execute DNN inference tasks on GPUs due to their strong computing capacity. Although studies have been conducted to profile the inference time of different DNNs on mobile GPUs and optimize it \cite{Jiang10}, the model reflecting the influence of GPU frequency on DNN inference time is not given. In \cite{Shi11}, the authors formulate the DNN inference time as the ratio of required computation workload to the computing frequency, but details on how to get the computation workload on GPUs are not given. Different from previous works, the authors in \cite{Zeng12} employ a model where the DNN inference time on GPUs is represented by the ratio between the number of floating-point operations (FLOPs) needed for DNN inference and the computing speed of GPUs (i.e., the product of GPU frequency and the number of FLOPs per cycle). 

Although these existing works formulate DNN inference time as the ratio of required workload to computing frequency, they do not provide realistic experiment data to verify the relationship. According to the CPU-DVFS model, DNN inference time is proportional to its required FLOPs when GPU frequency is fixed. To this end, we deploy five widely-used DNNs on NVIDIA Jetson Xaiver NX and Orin Nano to obtain the average inference time with the GPU frequency fixed, as shown in Fig. \ref{fig:myFig1}. When fitting the relationship between GPU inference time and required FLOPs as a linear function, the correlation coefficient shows a weak relation. For example, VGG19 \cite{Simonyan13} requires more FLOPs than ResNet152 \cite{He14}, but the GPU inference time of VGG19 on Jetson Xavier NX or Jetson Orin Nano is smaller than ResNet152. The real-world data further presents the inconsistency between the required computation workload and actual inference time. Therefore, directly applying the CPU-DVFS model to DNN inference on GPUs gives rise to distinct errors and is often not applicable. 

In this paper, we characterize a realistic model to show the impact of GPU frequency on DNN inference time based on the DVFS technique. To assist partial offloading, we partition DNNs into multiple blocks and obtain the actual inference time of different blocks to further verify the proposed model. We mainly focus on reducing the energy consumption by our proposed model and DVFS technique. Specifically, we consider the local inference and device-edge cooperative inference scenarios to evaluate the proposed model. Evaluation results show that in the local inference case, optimizing with our proposed model can significantly reduce the latency and energy consumption of devices under the corresponding resource constraints. In the scenario of cooperative inference, the partition policy optimized using our proposed model can achieve the minimum energy consumption under the deadline constraint. 

\section{System Overview}\label{sec:system}
\subsection{System Model}

\begin{figure}[t]
  \centering
  \includegraphics[width=0.3\textwidth]{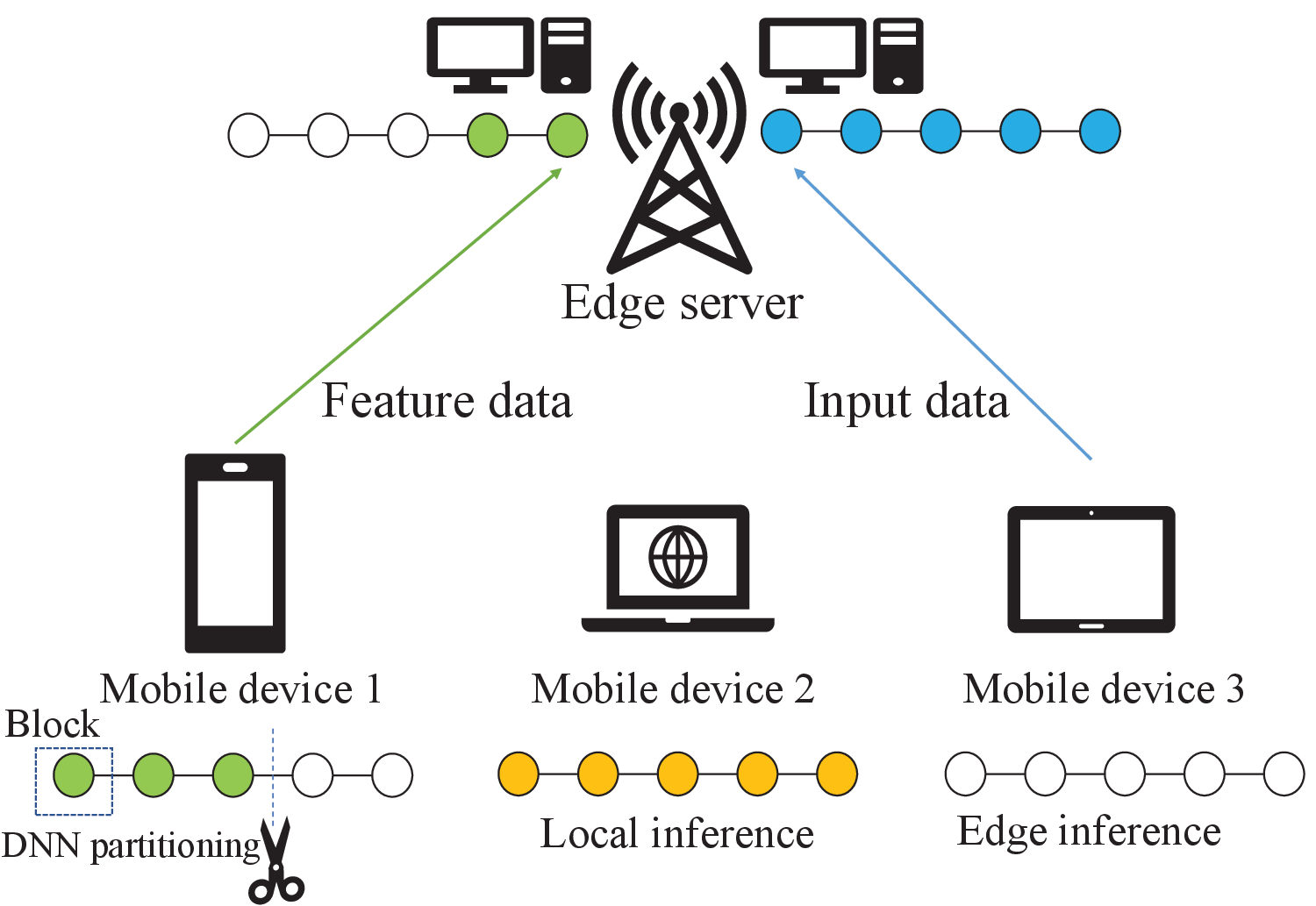}\\
  \caption{\leftline{Illustration of an edge intelligence system.}}
  \label{fig:myFig2}
  \vspace{-0.25cm}
\end{figure}

As shown in Fig. \ref{fig:myFig2}, we consider an edge intelligence system consisting of $N$ mobile devices, denoted as $\mathcal{N} \triangleq \{1, 2, \dots, N\}$, and an edge server. Each device needs to execute inference tasks and deploys common DNNs (e.g., AlexNet \cite{Kriz15}, VGG or ResNet). By combining multiple layers (e.g., convolution, pooling, activation) into a block, DNNs can be modeled as a sequence of $M$ blocks, which is a serial structure and easy to partition. The partitioning point set is denoted as $\mathcal{M} \triangleq \{0, 1, \dots, M\}$. For example, in Fig. \ref{fig:myFig2} the DNN consists of $M = 5$ blocks, and mobile device 1 partitions the DNN at point $m = 3$. The partition point of mobile device 2 is $m = 5$, which means it executes local inference. Mobile device 3 partitions DNN at point $m = 0$, offloading the input data to the edge server and executing edge inference. For simplicity, we assume that the downloading time of the inference result is ignored due to its much smaller data size.

\subsection{Inference Time and Energy Consumption}
\begin{figure}[t]
  \centering
  \includegraphics[width=0.32\textwidth]{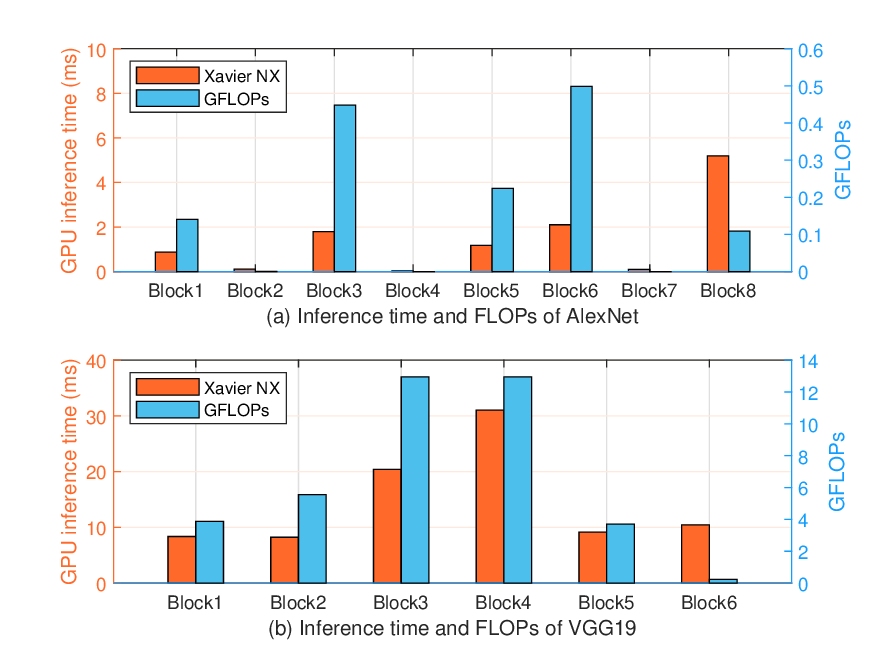}\\
  \caption{Average GPU inference time and FLOPs of different blocks for (a) AlexNet and (b) VGG19 on Jetson Xavier NX.}
  \label{fig:myFig3}
  \vspace{-0.25cm}
\end{figure}

The CPU-DVFS model formulates DNN inference time as
\begin{align}\label{eq:myeq1}
    t_{n} = \frac{w_n}{g_n f_n}, \forall n \in \mathcal{N},
\end{align}
where $w_n$, $g_n$ and $f_n$ denote the computation workload (in FLOPs), the number of FLOPs per cycle, and the GPU frequency of device $n$ (in cycle/s), respectively \cite{Shi11}, \cite{Zeng12}. However, the formula is not realistic because Fig. \ref{fig:myFig1} shows that the total inference time of a DNN is not proportional to its required FLOPs when GPU frequency is fixed. To further present the inconsistent relationship between inference time and FLOPs, we partition AlexNet and VGG19 on Jetson Xaiver NX, and the results are shown in Fig. \ref{fig:myFig3}. We observe that the required FLOPs of block 6 in AlexNet is larger than that of block 8, but the actual inference time of block 6 is smaller. The needed FLOPs of block 6 in VGG19 is much smaller than block 5, but their inference time is very close. These results further show that the CPU-DVFS model is not always applicable for different DNN blocks. Using this model to formulate DNN inference time on GPUs gives an inaccurate estimation. Therefore, we characterize a more precise inference time model to refine the CPU-DVFS model, and more details are given in Section \ref{sec:model}.

The power consumption of CMOS circuit consists of static and dynamic parts. The static power consumption remains constant when frequency is scaled, so we consider the dynamic part. The dynamic power of CMOS circuit is usually formulated as $P = \alpha C V^2 f$, where $\alpha$ is the activity factor, $C$ is the load capacitance, $V$ is the voltage and $f$ is the clock frequency \cite{Haj16}. By the DVFS technique, the frequency can be adjusted and available scales are usually in linear zones, which means that $V \propto f$. Therefore, the energy consumption of local inference on device $n$ is 
\begin{align}\label{eq:myeq2}
    e_{n} = \kappa_n f_n^3 t_{n}, \forall n \in \mathcal{N},
\end{align}
where $\kappa_n$ is an equivalent coefficient \cite{Liu17}.

\section{Modeling and Analysis of DNN Inference Time}\label{sec:model}
\subsection{DNN Inference Time}
\begin{figure}[t]
  \centering
  \includegraphics[width=0.35\textwidth]{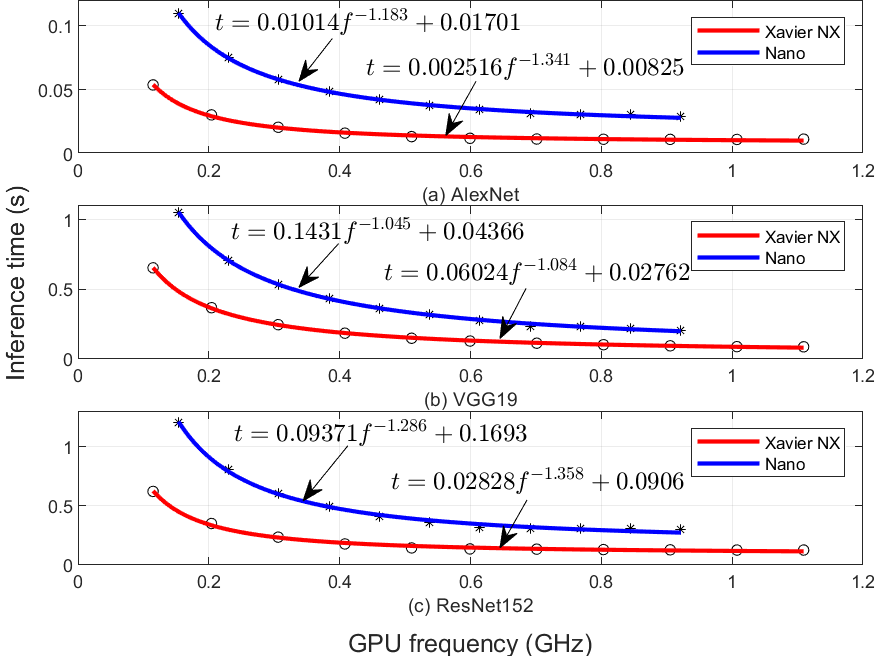}\\
  \caption{Fitting models of GPU inference time for (a) AlexNet, (b) VGG19 and (c) ResNet152 on Jetson Xavier NX and Jetson Nano.}
  \label{fig:myFig4}
  \vspace{-0.3cm}
\end{figure}
\begin{figure}[t]
  \centering
  \includegraphics[width=0.32\textwidth]{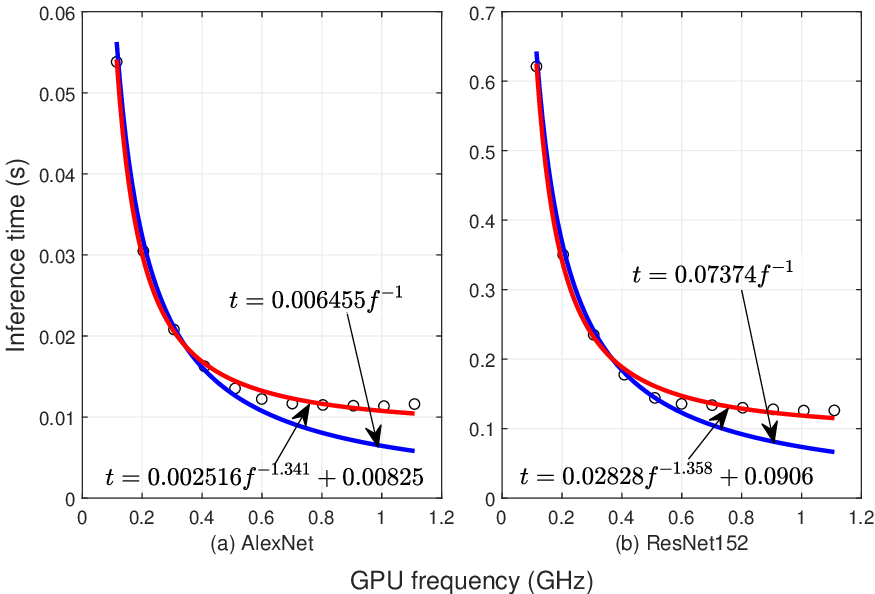}\\
  \caption{Comparison between the proposed model and the CPU-DVFS model for (a) AlexNet and (b) ResNet152 on Jetson Xavier NX.}
  \label{fig:myFig5}
  \vspace{-0.3cm}
\end{figure}

The computing capacity of a GPU is typically described by the maximum number of FLOPs it can execute per second (i.e., the peak computing capacity). However, the ratio of required FLOPs to the peak computing capacity cannot reflect the impact of GPU frequency on inference time. Aforementioned results also show that the actual GPU inference time is not proportional to required FLOPs with GPU frequency unchanged. To obtain a realistic model, we change the GPU frequency and measure the average inference time of three different DNNs on Jetson Xavier NX and Jetson Nano, as shown in Fig. \ref{fig:myFig4}. Based on the fitting results, the DNN inference time is formulated as
\begin{align}\label{eq:myeq3}
    t_n = a_n f^{-b_n} + c_n, \forall n \in \mathcal{N}, 
\end{align}
where $a_n, b_n, c_n > 0$ are coefficients related to numerous factors (e.g., the architecture and required workload of DNNs, the memory of device, etc.). First, we improve the inconsistent relationship between FLOPs and inference time. Parameter $a_n$ is similar to the effect of $\frac{w_n}{g_n}$, which mainly reflects the computing workload. Parameter $b_n$ reflects the impact of frequency, which is close but not always equal to 1. Using the CPU-DVFS model to fit data will give a wrong estimation of inference time, which is presented in Fig. \ref{fig:myFig5} in detail. Although the two models have similar trends in the low frequency range, they begin to separate when GPU frequency exceeds a certain threshold. The CPU-DVFS model no longer applies for AlexNet and ResNet152 when GPU frequency is larger than 0.6 GHz. The required FLOPs of AlexNet and ResNet152 are 1.43G and 23.11G, respectively. The ratio of their required FLOPs is $\frac{1.43}{23.11} = 0.062$, while the corresponding ratio of fitting coefficients adopting the CPU-DVFS model is $\frac{0.006455}{0.07374} = 0.088$. The realistic experiment results further show the flaws of the CPU-DVFS model in characterizing DNN inference time on GPUs.

Next, different from the CPU-DVFS model, we add a constant term $c_n$ to better formulate the DNN inference time. That is due to the fact that when GPU frequency is larger than a threshold, the inference time will not dramatically decrease. This phenomenon shows that the impact of GPU frequency on inference time is not obvious in the high frequency range, since the memory access speed plays a leading role in the DNN inference time. In summary, we use parameters $a_n$ and $b_n$ to represent the effect of computing workload and frequency on inference time, while the other possible factors are characterized by the parameter $c_n$.

\begin{table*}[htbp]
    \centering
    \caption{Fitting parameters for different blocks of AlexNet on Jetson Xavier NX.}
    \begin{tabular}{c|cccccccc}
    \hline
    Parameters & Block1 & Block2 & Block3 & Block4 & Block5 & Block6 & Block7 & Block8 \\
    \hline
    $a_{n,m}$ & 0.7111 & 0.0339 & 0.2627 & 0.0239 & 0.7843 & 1.158  & 0.0553 & 0.8595\\
    \hline
    $b_{n,m}$ & 0.750  & 0.745  & 1.372  & 0.904  & 0.896  & 1.113  & 0.905  & 1.432 \\
    \hline
    $c_{n,m}$ & 0.0865 & 0.0295 & 1.601  & 0.0018 & 0.1163 & 0.8472 & 0.0065 & 3.843 \\
    \hline
    \end{tabular}
    \label{tab:mytab1}
    \vspace{-0.3cm}
\end{table*}

\begin{table*}[htbp]
    \centering
    \caption{Fitting parameters for different blocks of ResNet152 on Jetson Xavier NX.}
    \begin{tabular}{c|ccccccccc}
    \hline
    Parameters & Block1 & Block2 & Block3 & Block4 & Block5 & Block6 & Block7 & Block8 & Block9 \\
    \hline
    $a_{n,m}$ & 1.009 & 0.7454 & 2.196 & 1.153 & 5.288 & 4.533  & 4.141 & 5.544 & 5.85 \\
    \hline
    $b_{n,m}$ & 0.669 & 1.614  & 1.402  & 1.529  & 1.374  & 1.371  & 1.407  & 1.325 & 1.027 \\
    \hline
    $c_{n,m}$ & 0.2721 & 7.168 & 8.44  & 7.743 & 17.92 & 15.03 & 16.14 & 14.84 & 0.8289 \\
    \hline
    \end{tabular}
    \label{tab:mytab2}
    \vspace{-0.2cm}
\end{table*}

\begin{figure*}[t]
  \centering
  \includegraphics[width=1.0\textwidth]{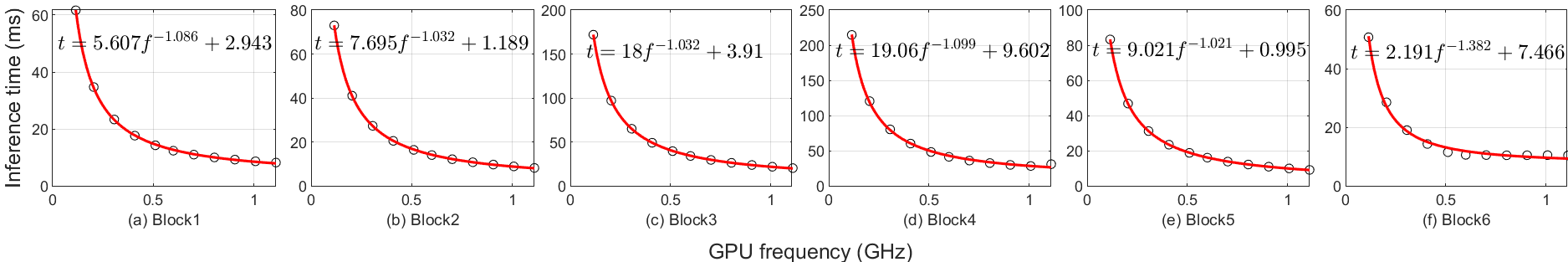}\\
  \caption{\leftline{Fitting models of GPU inference time for different blocks of VGG19 on Jetson Xavier NX.}}
  \label{fig:myFig6}
  \vspace{-0.2cm}
\end{figure*}

\subsection{Inference Time of DNN Blocks}
\begin{figure}[t]
  \centering
  \includegraphics[width=0.3\textwidth]{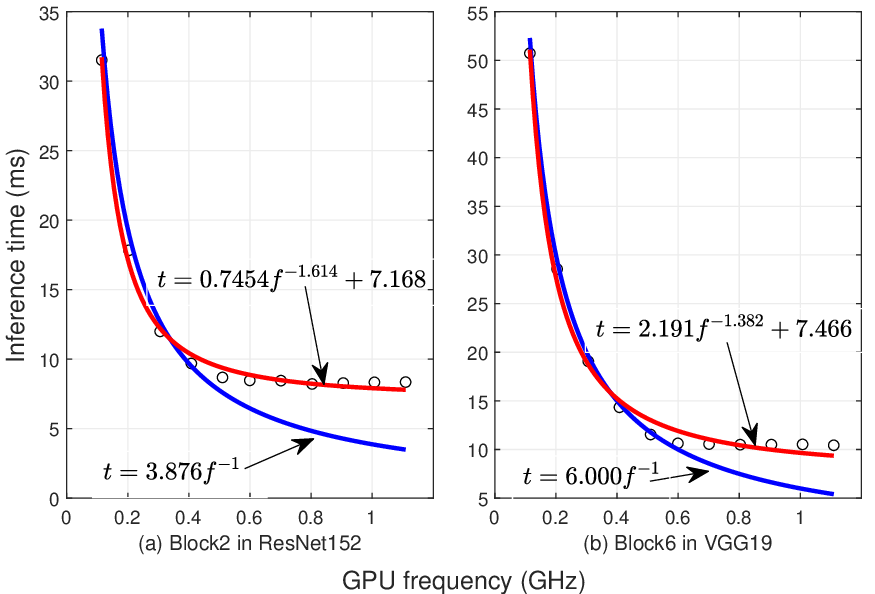}\\
  \caption{Comparison between the proposed model and the CPU-DVFS model for specific blocks.}
  \label{fig:myFig7}
  \vspace{-0.3cm}
\end{figure}

DNN partitioning is a usual technique in cooperative inference, and modeling the inference time of different blocks is essential to obtain effective partition policies. To this end, we adjust the GPU frequency and measure the inference time of different blocks in AlexNet, VGG19 and ResNet152 on Jetson Xavier NX and Nano. We just present the fitting curves of VGG19 on Jetson Xavier NX, as shown in Fig. \ref{fig:myFig6}. In view of these results, the GPU inference time of the $m$th block for device $n$ is given as
\begin{align}\label{eq:myeq4}
    t_{n,m} = a_{n,m} f^{-b_{n,m}} + c_{n,m}, \forall n \in \mathcal{N}, m \in \mathcal{M}, 
\end{align}
where $a_{n,m}, b_{n,m}$ and $c_{n,m} > 0$ are not only relevant to the factors that affect the total inference time but also the partitioning method. The corresponding parameters for AlexNet and ResNet152 on Jetson Xavier NX are presented in Table \ref{tab:mytab1} and Table \ref{tab:mytab2}, respectively. The inference time is in ms and the GPU frequency is in GHz. Based on the results, we observe that the proposed model also applies for the inference time of each block. Similarly, when GPU frequency is larger than a threshold, the inference time of different blocks will decline more slowly, or even keep constant. We compare our proposed model with the CPU-DVFS model for two specific blocks in Fig. \ref{fig:myFig7}. Fitting inference time according to the CPU-DVFS model will overestimate the computing capacity of GPUs in high frequency scales, while our proposed model can better formulate the actual time. We can also provide appropriate GPU frequency ranges to optimize the inference time and energy consumption of mobile devices.

\section{Evaluation Results}\label{sec:eval}

\subsection{Evaluation of Local Inference}
We use two classical DNNs (i.e., VGG19 and ResNet152) to execute the image classification task on CIFAR10 dataset \cite{Kriz18}. The device is chosen as NVIDIA Jetson Xavier NX and the available GPU frequency range is $f \in [0.12, 1.10]$ GHz. We fix the GPU frequency and measure the GPU power consumption, then the equivalent energy coefficient $\kappa_n = 1.3 \ \mathrm{W/(Hz)}^3$ can be computed based on e.q. (\ref{eq:myeq2}). We choose the CPU-DVFS model as the benchmark, where the number of FLOPs per cycle $g_n$ is set as 1536 \cite{Dol19}. 

\begin{figure}[t]
  \centering
  \includegraphics[width=0.3\textwidth]{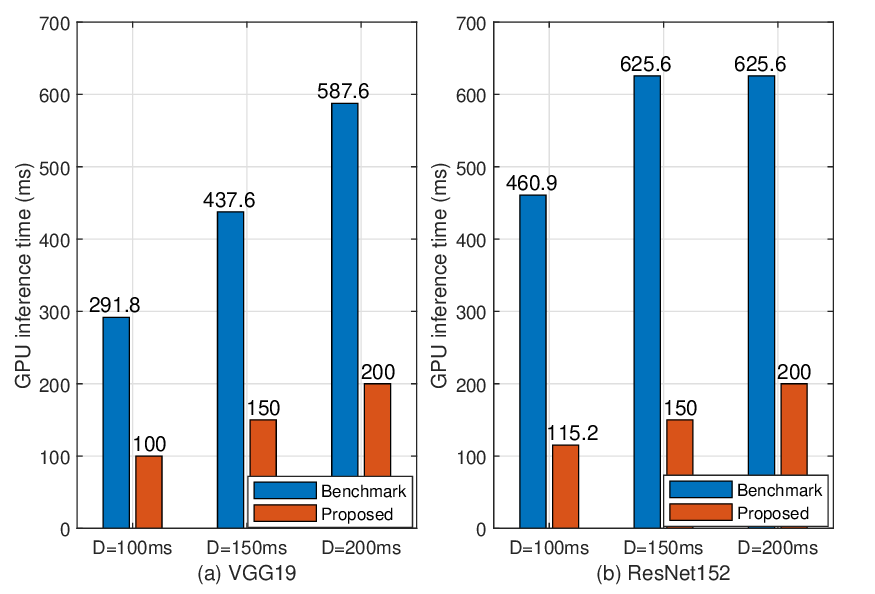}\\
  \caption{GPU inference time comparison between using the proposed and the benchmark model for (a) VGG19 and (b) ResNet152 with different execution deadlines.}
  \label{fig:myFig8}
  \vspace{-0.3cm}
\end{figure}

In Fig. \ref{fig:myFig8}, we compare the actual inference time based on these two different models under three given execution time deadlines (i.e., $D = 100$ ms, 150 ms, and 200 ms). The optimal is to adjust the GPU frequency to make the inference tasks completed within the given deadlines and reduce the energy consumption as much as possible. It is observed that optimization with our proposed model can give smaller inference time compared with the benchmark. The reason is that the benchmark model underestimates the DNN inference time, which leads to smaller GPU frequency and makes the inference tasks unable to be completed within the given deadline in practice. In addition, even if we set the GPU frequency to the maximum for ResNet152 under the deadline $D = 100$ ms, the actual inference time is also larger than the given deadline due to the limited GPU frequency scale. When the deadline is set as 150 ms and 200 ms, according to the benchmark model, using the minimum GPU frequency can meet the constraint and save energy. Therefore, the benchmark model gives the same inference time in these two situations, which is much larger than the given deadlines. For VGG19, using our proposed model will give the same inference time as the deadline because the execution deadline can be achieved at an available frequency. Similarly, the benchmark model results in smaller GPU frequency and larger inference time. 

\begin{figure}[t]
  \centering
  \includegraphics[width=0.3\textwidth]{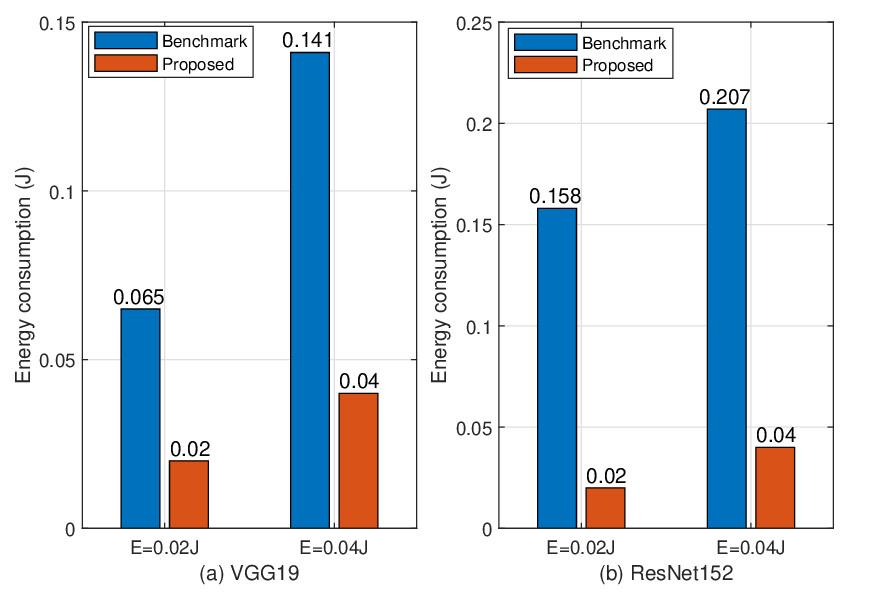}\\
  \caption{GPU energy consumption comparison between using the proposed and the benchmark model for (a) VGG19 and (b) ResNet152 with different energy constraints.}
  \label{fig:myFig9}
  \vspace{-0.3cm}
\end{figure}

In Fig. \ref{fig:myFig9}, we evaluate the realistic energy consumption based on the proposed and benchmark model under two different energy constraints (i.e., $E = 0.02$ J and 0.04 J). Obviously, the inference task can be completed within the given energy constraint by our proposed model, while the benchmark model brings larger energy consumption. It is because the benchmark model underestimates the energy consumption. With a given energy constraint, the benchmark model will set a larger GPU frequency and result in greater energy consumption. For VGG19 and ResNet152, the proposed model can reduce the energy consumption by about 70\% and 84\% respectively compared with the benchmark. In conclusion, the proposed model can provide more realistic guidance for adjusting GPU frequency to optimize latency and energy consumption.

\subsection{Evaluation of a special case for cooperative inference}
\begin{figure}[t]
  \centering
  \includegraphics[width=0.3\textwidth]{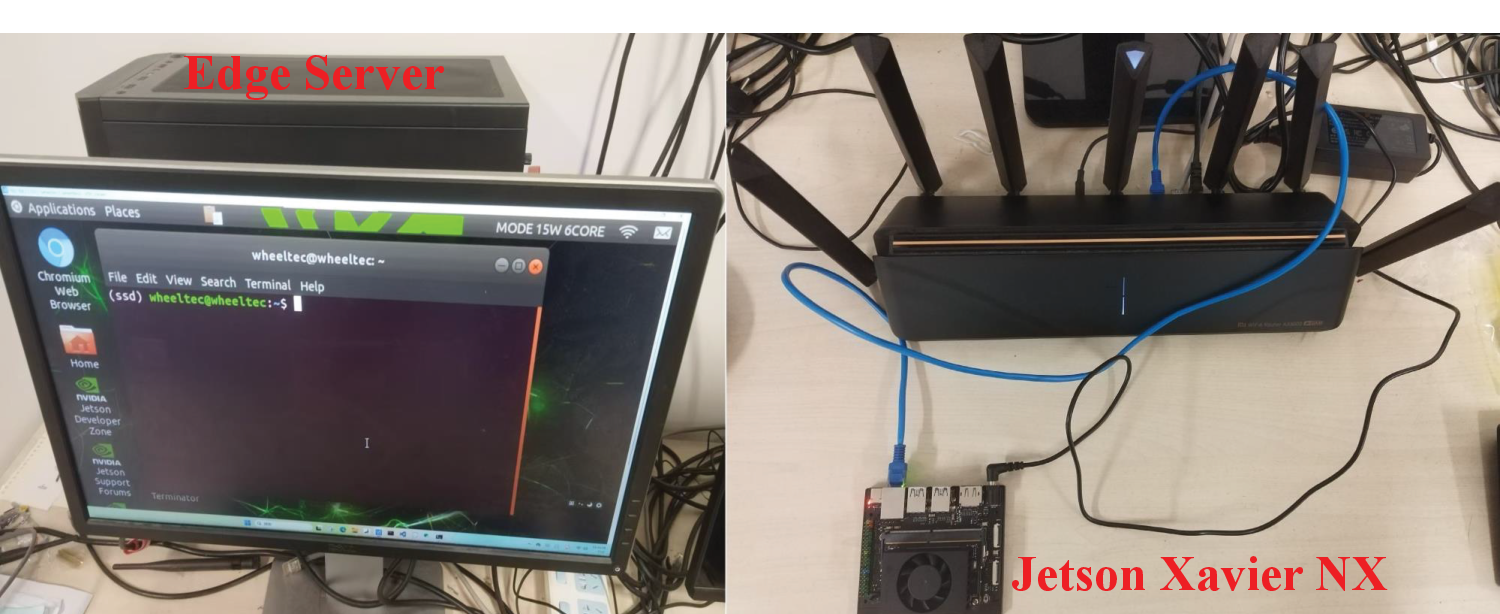}\\
  \caption{\leftline{Testing environment for cooperative inference.}}
  \label{fig:myFig10}
  \vspace{-0.3cm}
\end{figure}

We consider a special cooperative inference scenario to verify the proposed model, where the system consists of one device and an edge server. The device is selected as Jetson Xavier NX and the edge server is GeForce RTX 4080, as shown in Fig .\ref{fig:myFig10}. The inference tasks (e.g., image classification) should be processed within a given deadline. By offloading part of DNN execution to the edge server, the computation intensity of mobile devices can be relieved and the total latency can be reduced to meet the deadline requirement. We use ResNet152 and VGG19 to execute inference tasks, which are divided into 9 and 6 blocks respectively. The input image is 0.57 MB, and the deadline is set as $D = 200$ ms. The output feature size of different blocks in ResNet152 and VGG19 are $\{3.06, 0.77, 1.53, 0.38, 0.19, 0.19, 0.19, 0.10, 3.8\times10^{-5}\}$ MB and $\{3.06, 1.53, 0.77, 0.38, 0.10, 3.8\times10^{-5}\}$ MB, respectively. The objective is to select a partition point to minimize the summation of computation and communication energy, while the inference task can be completed within the given deadline. For simplicity, we just fix the communication rate so that the effect of inference time models can be obvious.

\begin{figure}[t]
  \centering
  \includegraphics[width=0.35\textwidth]{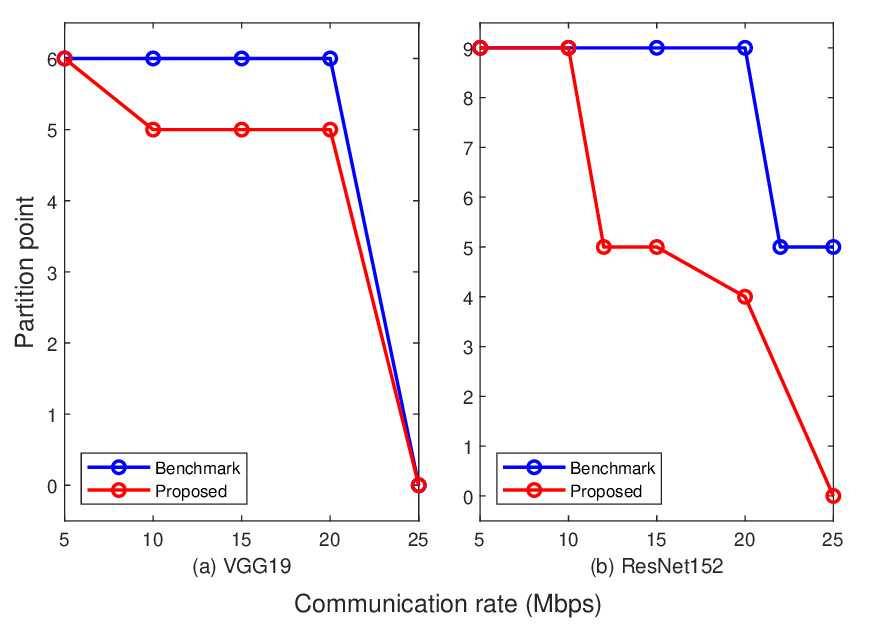}\\
  \caption{Partition points of (a) VGG19 and (b) ResNet152 given by the benchmark and proposed model under different communication rates.}
  \label{fig:myFig11}
  \vspace{-0.3cm}
\end{figure}

In Fig. \ref{fig:myFig11}, we compare our proposed model with the benchmark model in terms of the partition policy under different communication rates. When communication rate is relatively small, the partition policy shows that the device chooses to execute local inference. With larger communication rates, the device will partition at an intermediate point and offload the rest of DNNs to the edge server. It is obvious that the partition policy given by the benchmark model tends to make the device execute inference locally when the communication rate is smaller than 20 Mbps, since the benchmark model underestimates the local inference time. However, using our proposed model can provide refined partition policies for VGG19 and ResNet152 when communication rate is larger than 10 Mbps. In particular, when communication rate is 25 Mbps, these two models give the same partition policy (i.e., offloading the input data) for VGG19. This is because the smaller feature data is located in the latter part of VGG19, which brings large local inference cost.

\begin{figure}[t]
  \centering
  \includegraphics[width=0.3\textwidth]{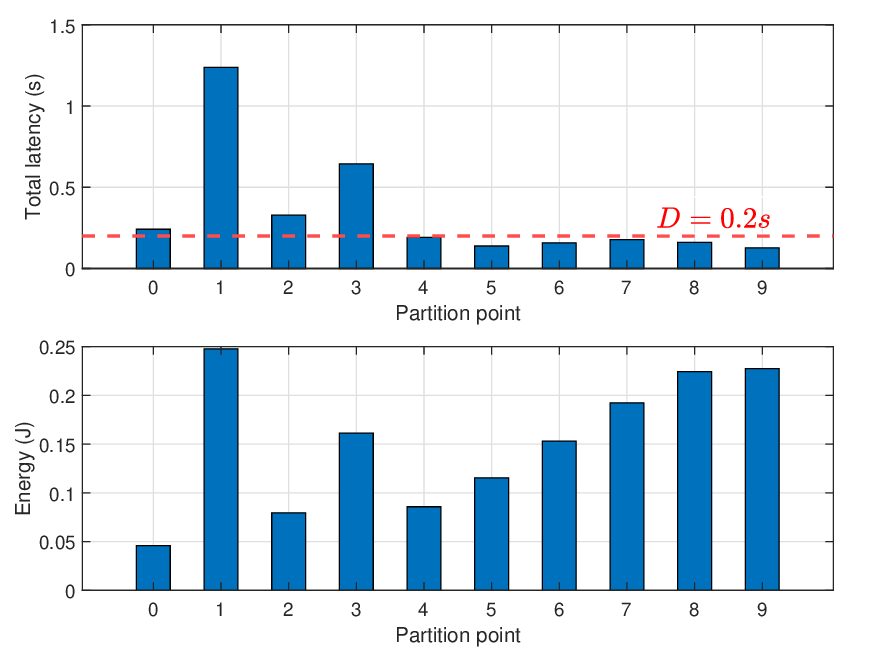}\\
  \caption{Total latency and energy for different partition points of ResNet152 with communication rate $R = 20$ Mbps.}
  \label{fig:myFig12}
  \vspace{-0.3cm}
\end{figure}

\begin{figure}[t]
  \centering
  \includegraphics[width=0.3\textwidth]{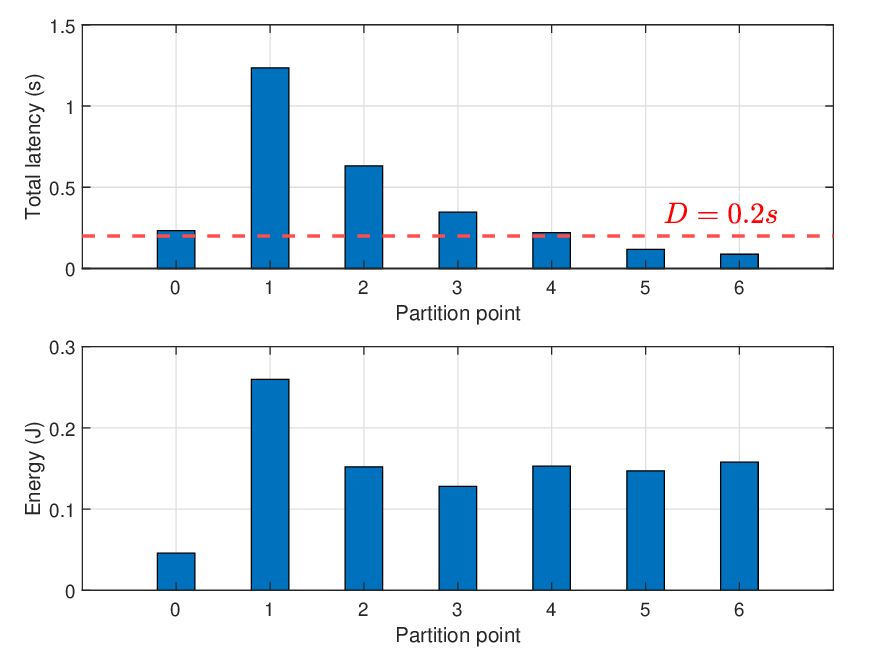}\\
  \caption{Total latency and energy for different partition points of VGG19 with communication rate $R = 20$ Mbps.}
  \label{fig:myFig13}
  \vspace{-0.3cm}
\end{figure}

To further analyze the impact of different partition policies, in Fig. \ref{fig:myFig12} and Fig. \ref{fig:myFig13} we investigate the total latency and energy consumption for different partition points of ResNet152 and VGG19 respectively when communication rate is fixed as 20 Mbps. From Fig. \ref{fig:myFig11}, we know that these two models yield different partition points. Using the benchmark model gives the 9th partition point in ResNet152, which means executing local inference, while our proposed model partitions ResNet152 at the 4th block. For ResNet152, it is observed that the partition policy given by our proposed model can achieve the minimum energy consumption with the deadline constraint. The energy consumption is reduced by 62.3\% compared with the benchmark when using ResNet152 to execute inference. According to our proposed model, we should partition VGG19 at the 5th block rather than 6th block based on the benchmark. However, the impact of partitioning VGG19 at the 5th or 6th point on the energy consumption is not highly prominent in this setting. This is because the communication cost of offloading feature data output by the 5th partition point is comparable to the computation cost of executing the 6th block locally. To sum up, our proposed inference time model can provide better guidance for partitioning policy design than the benchmark in multiple scenarios.

\section{Conclusion}\label{sec:conclusion}
In this paper, we propose a realistic model for the inference time of DNNs on GPUs based on the DVFS technique. We analyze the difference between the proposed model and the CPU-DVFS model from the perspective of fitting parameters, especially for the high frequency range. Experiment data for different blocks in DNNs further validates the proposed model. Evaluation results show that using our proposed model for optimization can reduce the latency and energy consumption by at least 66\% and 69\% compared with the CPU-DVFS model in local inference scenarios. For the cooperative inference case, the partition policy with our proposed model achieves an improvement in the energy consumption of mobile devices compared with the benchmark policy. For future work, more experiments on different mobile devices and DNNs should be conducted, and a realistic dataset will be constructed to provide relevant parameters for optimization.



\end{document}